\documentclass[conference]{IEEEtran}
\IEEEoverridecommandlockouts
\newtheorem{definition}{Definition}
\usepackage{cite}
\usepackage{amsmath,amssymb,amsfonts}
\usepackage{algorithmic}
\usepackage{graphicx}
\usepackage{textcomp}
\usepackage{xcolor}
\def\BibTeX{{\rm B\kern-.05em{\sc i\kern-.025em b}\kern-.08em
    T\kern-.1667em\lower.7ex\hbox{E}\kern-.125emX}}

\usepackage{subfig} 

\usepackage{mathtools}
\usepackage[utf8]{inputenc}
\usepackage[english]{babel}
\usepackage[numbers]{natbib}

\usepackage[export]{adjustbox}
\usepackage{algorithm,algorithmic}
\usepackage{hyperref}
\usepackage{enumitem}
\usepackage{textcomp}
\usepackage{xcolor}
\usepackage{comment}
\usepackage{subfig}
\usepackage{adjustbox}
\usepackage{array}
\usepackage{booktabs}
\usepackage{multirow}
\usepackage{hhline}
\usepackage{mathtools}
\usepackage{dsfont}

\usepackage{times}
\usepackage{soul}
\usepackage{url}
\usepackage[utf8]{inputenc}
\usepackage{booktabs}
\DeclareMathOperator*{\argmin}{arg\,min}
\DeclareMathOperator*{\argmax}{arg\,max}
\usepackage{times}
\usepackage{helvet}
\usepackage{courier}
\usepackage[numbers]{natbib}
\usepackage{amssymb,amsfonts}
\usepackage{soul}
\usepackage{url}
\usepackage[utf8]{inputenc}
\usepackage{graphicx}
\usepackage{amsmath}
\usepackage{booktabs}
\usepackage[utf8]{inputenc}
\usepackage[english]{babel}
\usepackage{amsmath}
\usepackage{times}
\usepackage{helvet}
\usepackage{courier}
\usepackage[numbers]{natbib}

\usepackage{lipsum}
\usepackage[export]{adjustbox}
\usepackage{algorithm,algorithmic}
\usepackage{hyperref}
\usepackage{enumitem}
\usepackage{graphicx}
\usepackage{textcomp}
\usepackage{bbm}
\usepackage[utf8]{inputenc}
\usepackage{xcolor}
\usepackage{comment}

\usepackage[export]{adjustbox}
\usepackage{algorithm,algorithmic}
\usepackage[utf8]{inputenc}
\usepackage{lipsum}

\newcommand\blfootnote[1]{%
  \begingroup
  \renewcommand\thefootnote{}\footnote{#1}%
  \addtocounter{footnote}{-1}%
  \endgroup
}

\begin{document}

\title{
Adversarial Learning for Counterfactual Fairness } 

\author{\IEEEauthorblockN{Vincent Grari}
\IEEEauthorblockA{\textit{Sorbonne Université
LIP6/CNRS}\\
Paris, France \\
vincent.grari@lip6.fr}
\and
\IEEEauthorblockN{Sylvain Lamprier}
\IEEEauthorblockA{\textit{Sorbonne Université
LIP6/CNRS}\\
Paris, France \\
sylvain.lamprier@lip6.fr}
\and
\IEEEauthorblockN{Marcin Detyniecki}
\IEEEauthorblockA{\textit{AXA
REV Research}\\
Paris, France \\
marcin.detyniecki@axa.com}
}

\maketitle

\begin{abstract}

In recent years, fairness has become an important topic in the machine learning research community. In particular,  counterfactual fairness aims at building prediction models which ensure fairness at the most individual level. Rather than globally considering equity over the  entire population, the idea is to imagine what any individual would look like with a variation of a given attribute of interest, such as  a different gender or race for instance. 
Existing approaches rely on Variational Auto-encoding of individuals, 
using Maximum Mean Discrepancy (MMD) penalization 
to limit the statistical dependence of  inferred representations with 
their corresponding sensitive attributes. 
This enables the simulation of counterfactual samples used for training the target fair model,  
the goal being to produce similar outcomes for every alternate version of any individual.   
In this work, 
we propose to  rely on  
an adversarial neural learning approach, that enables more powerful inference than with MMD penalties, and is  particularly better fitted  for the continuous setting, where values of sensitive attributes cannot be exhaustively enumerated.  
Experiments show significant improvements 
in term of counterfactual fairness 
for both the discrete and the  continuous settings.


\begin{IEEEkeywords}
Counterfactual Fairness, Adversarial Neural Network, Causal Inference
\end{IEEEkeywords}
\end{abstract}
\section{Introduction}

\blfootnote{A link to the online repository will be provided upon acceptance} 

Machine learning models have an increasingly important role in our daily lives and can have significant implications for citizens like loan applications, recidivism score, credit rating, etc. However, the data used for training the models can reflect sensitive biases that exist in our society 
and without a careful design 
the models can perpetuate
or even reinforce these biases \citep{bolukbasi2016man}. Many incidents of this kind have been reported in recent years.
An infamous example is the case of a tool for criminal risk prediction  
(COMPAS), which showed strong discrimination against black defendants \citep{angwin2016machine}.

A fair predictive model 
 provides outcomes that do not contain any prejudice or favoritism toward an individual or a group  based on a set of sensitive characteristics. One of the problems in achieving a non-discriminatory model is that it is not simply a matter of removing protected attributes from the training base \citep{pedreshi2008discrimination}. This concept, known as \emph{fairness through unawareness}, is highly insufficient because any other non-sensitive attribute might indirectly contain significant  sensitive information. 
To tackle this problem the recent  \emph{fair machine learning} research field has emerged. 

As of now, a large majority of   
works in the field 
focused on  
group fairness metrics, that assess a form of conditional independence between the 3 following features: the sensitive attribute $A$, the true outcome feature $Y$, and the output model predictions $\hat{Y}$. For example, one of the most known objective \emph{Demographic parity} ensures that the output prediction is not dependent of the sensitive feature \citep{calders2009building,zafar2015fairness}. However, predictive models trained to be fair regarding  such  group metrics may  induce dramatic consequences for some individuals. For example in an  extreme case, a person may be refused a position only because of belonging to a privileged group, regardless of their merit within the group.
To tackle such issues, a recent field called Counterfactual fairness \citep{kusner2017counterfactual} 
recently proposed  to assess fairness at the individual level, by leveraging causal inference to ensure that some 
sensitive attributes are not the cause of a prediction change. 
It argues to lead to a more intuitive, powerful, and less error-prone way of reasoning about fairness \citep{chiappa2019path}. 
The idea is to imagine what any individual would look like with a variation of a given attribute of interest, such as  a different gender or race for instances, in order to ensure similar outcomes  
for every alternate version of the same individual.
While plenty of 
methods have been proposed recently to tackle this challenge for discrete variables, 
to the best of our knowledge no approach address the continuous case. The existing approches may not hold when, for
instance, the sensitive attribute is the age or the weight of an individual.

The main contributions of this paper are: 
\begin{itemize}
    \item We propose an adversarial approach for confounding variable inference, which allows the generation of accurate counterfactuals in both discrete and continuous sensitive settings (while existing approaches are limited to the discrete case); 
    \item Based on this, we define an approach for counterfactual fairness tolerant to continuous features, notably via a dynamic sampling method that focuses on individualized hard locations of the sensitive space;   
    \item We demonstrate empirically that our algorithm can mitigate counterfactual fairness 
\end{itemize}
 
 
 Section 2 first gives details 
 for counterfactual fairness, which we believe are essential for a good understanding of our contributions. Then, in section \ref{model}, we detail our approach in two main steps. Section \ref{xp} evaluates  performances for both the discrete and the continuous settings.

 \section{Background}
\label{bg}


Recently, there has been a dramatic rise of interest for fair machine learning by the academic community. Many questions have been raised, such as:  How to define fairness \citep{hinnefeld2018evaluating,hardt2016equality,dwork2012fairness,kusner2017counterfactual} ? How to mitigate the sensitive bias  \citep{zhang2018mitigating,abs-1911-04929,kamiran2012data,bellamy2018ai,calmon2017optimized,zafar2015fairness,celis2019classification,wadsworth2018achieving,louppe2017learning,chen2019fairness,kearns2017preventing} ? How to keep a high prediction accuracy while remaining fair in a complex real-world scenario  \citep{DBLP:conf/icdm/GrariRLD19,adel2019one} ? To answer these questions, three main families  of fairness approaches exist in the literature. While   
pre-processing \citep{kamiran2012data,bellamy2018ai,calmon2017optimized} and post-processing  \citep{hardt2016equality,chen2019fairness} approaches respectively  act on the input or the output of a classically trained predictor,  
pre-processing \citep{kamiran2012data,bellamy2018ai,calmon2017optimized} and post-processing  \citep{hardt2016equality,chen2019fairness} approaches respectively  act on the input or the output of a classically trained predictor,  
in-processing approaches mitigate the undesired bias directly during the training phase \citep{zafar2015fairness,celis2019classification, zhang2018mitigating,wadsworth2018achieving,louppe2017learning}. In this paper we focus on in-processing fairness, which reveals as the most powerful framework for settings where acting on the training process is an option. 

Throughout this document, the aim is to learn a predictive function $h_\theta$ from training data that consists of $m$ examples ${(x_{i},a_{i},y_{i})}_{i=1}^{m}$, where $x_{i} \in \mathbb{R}^{p}$ is the $p$-sized feature vector $X$ of the $i$-th example, $a_i \in \Omega_{A}$ the value of its sensitive attribute 
and $y_{i}$ its label to be predicted. According to the setting, the domain $\Omega_{A}$ of the sensitive attribute $A$ can be either a discrete  or a continuous set. The outcome $Y$ is also either binary or continuous. 
The objective is to ensure some individual fairness guarantees on the outcomes of the predictor $\hat{Y}=h_\theta(X,A)$, 
by the way of Counterfactual Fairness. The remaining of this section presents classical and Counterfactual Fairness Metrics and existing methods for Counterfactual Fairness.

\subsection{Fairness definitions and metrics}

The vast majority of fairness research works have focused on 
two metrics that have become very popular in the fairness field: \emph{Demographic parity} \citep{dwork2012fairness} and \emph{Equalized odds} \citep{hardt2016equality}. 
Both of them consider 
fairness globally, by focusing on equity between groups of people, defined according to one or several high level sensitive attributes. The \emph{Demographic parity} metric  compares the average prediction for each demographic sensitive group. For instance, in the binary discrete case, it comes down to ensure that: $P(\widehat{Y}=1|A=0)=P(\widehat{Y}=1|A=1)$.
The underlying idea is that each sensitive demographic group must own the same chance for a positive outcome. The \emph{Equalized odds} metric rather compares rates of True positives and False positives between sensitive groups: $P(\widehat{Y}=1|A=0,Y=y)=P(\widehat{Y}=1|A=1,Y=y), \forall y\in\{0,1\}$.   The notion of fairness here is that chances for being correctly or incorrectly classified as positive should be equal for every group. However, these metrics which correspond to  averages over each sensitive groups are known to lead to  arbitrary individual-level fairness deviations, with a high outcome variance within  groups \citep{kearns2017preventing}. 

In the continuous setting, some recent works proposed to consider non-linear correlation metrics between the predicted outcome $\hat{Y}$ and the sensitive attribute $A$, such as the Hirschfeld-Gebelein-R\'enyi maximal correlation (HGR) defined, for two jointly distributed random variables $U \in \mathcal{U}$ and $V \in \mathcal{V}$,   as:
\begin{eqnarray}
HGR(U, V) = \sup_{\substack{ f:\mathcal{U}\rightarrow \mathbb{R},g:\mathcal{V}\rightarrow \mathbb{R}\\
           E(f(U))=E(g(V))=0 \\
           E(f^2(U))=E(g^2(V))=1}} \rho(f(U), g(V)) 
\label{hgr}
\end{eqnarray}
where $\rho$ is the Pearson linear correlation coefficient with some measurable functions $f$ and $g$. The HGR coefficient is equal to 0 if the two random variables are independent. If they are strictly dependent the value is 1. Applied to fairness \cite{pmlr-v97-mary19a,abs-1911-04929}, it can be used to measure objectives similar to those defined for the discrete setting, such as the \emph{Demographic parity}, which can be measured via $HGR(\widehat{Y},A)$ (this accounts for the  violation level of the constraint $P(\widehat{Y}|A)=P(\widehat{Y})$). 

However, even such approaches in the continuous setting only consider fairness globally and can lead to  particularly unfair decisions at the individual level.  For example, a fair algorithm can choose to accept a high MSE error for the outcome of a given person if this allows the distribution $P(\widehat{Y}|A)$ to get closer to $P(\widehat{Y})$. Penalization can be arbitrarily high on a given kind of individual profile compared to any other equivalent one, only depending on where 
the learning process converged. 
Global fairness is unfair. 

To tackle this problem,  Counterfactual fairness   has been recently introduced for quantifying fairness at the most individual sense \citep{kusner2017counterfactual}. The idea is 
to consider that a decision is fair for an individual if it coincides with the one that would have been taken in a counterfactual world in which the values of its sensitive attributes were different. It leverages the previous work  \citep{pearl2009causal}, which introduced a causal framework to learn from biased data by exploring the relationship between sensitive features and data. With the recent development in deep learning, some novels approaches \citep{madras2019fairness,pfohl2019counterfactual,louizos2017causal} argue to lead to a less error-prone decision-making model, by improving the approximation of the causal inference in the presence of unobserved confounders. 

\begin{definition}{Counterfactual demographic  parity  \citep{kusner2017counterfactual}}:
A predictive function $h_\theta$ is considered counterfactually fair for a causal world $G$, if for any $x \in X$ and $\forall y \in Y$,$\forall (a,a\prime) \in \Omega_{A}^2$ with $a \neq a\prime$:
$$p(\hat{Y}_{A\leftarrow{a}} =y|X=x,A=a) =p(\hat{Y}_{A\leftarrow{a\prime}} =y|X=x,A=a)
$$
where $\hat{Y}_{A\leftarrow{a'}} = h_\theta(\hat{X}_{A\leftarrow{a'}},a')$ is the outcome of the predictive function $h_\theta$ for any transformation $\hat{X}_{A\leftarrow{a'}}$ of input X, resulting from setting $a'$ as its sensitive attribute value, according  to the causal graph $G$.   
\label{def1}
\end{definition}
Following definition \ref{def1}, an algorithm is considered counterfactually fair in term of demographic parity if the predictions are equal for each individual in the factual causal world where $A=a$ and in any counterfactual world where $A=a^{\prime}$. It therefore compares  the predictions of the same individual with an alternate version of him/herself. Similar extension can be done to adapt the \textit{Equalized Odds} objective for  the Counterfactual framework \cite{pfohl2019counterfactual}. Learning transformations $\hat{X}_{A\leftarrow{a'}}$ for a given causal graph is at the heart of Counterfactual Fairness, as described in the next subsection.

\subsection{Counterfactual Fairness}
\label{counterfact}

\begin{figure}[h]
  \centering
  \includegraphics[scale=0.55,valign=t]{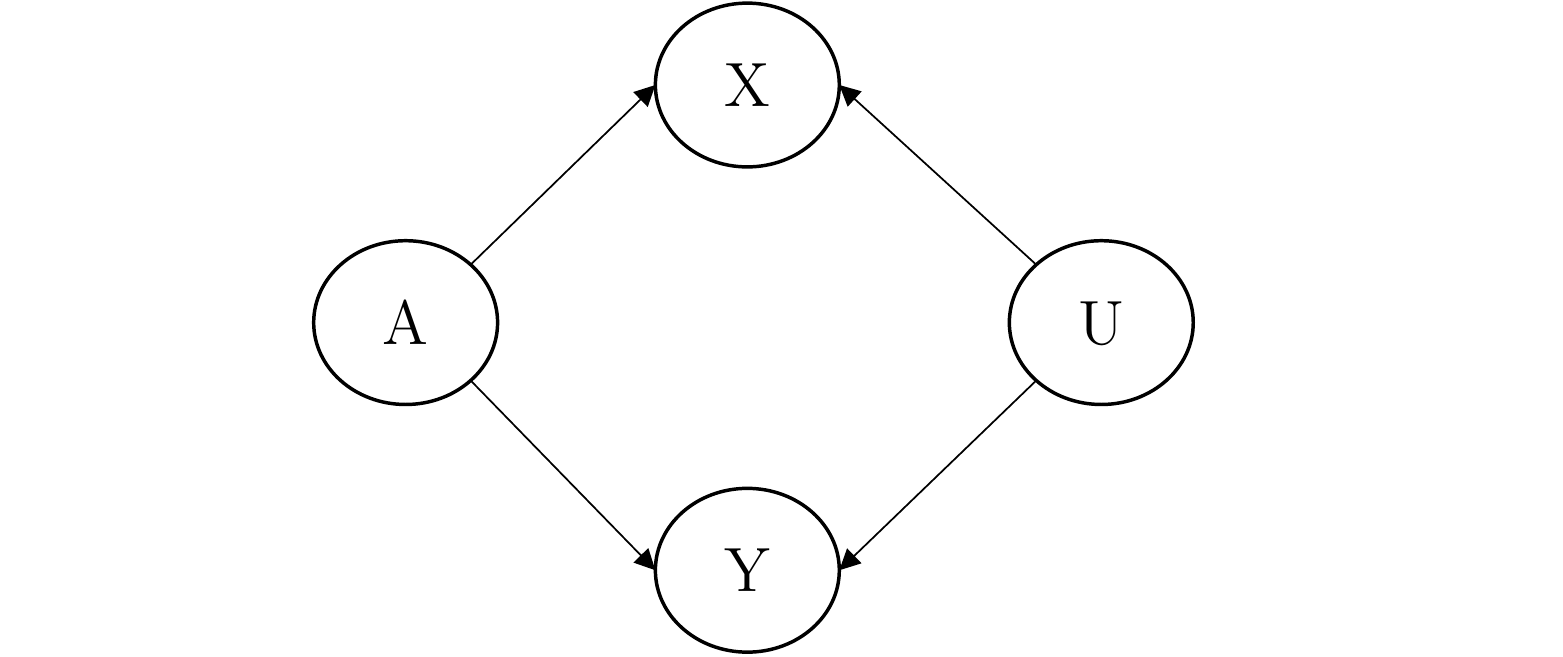}
  \caption{Graphical causal model. Unobserved confounder $U$ has effect on both $X$ and $Y$.}
  \label{fig:graph_causal}
\end{figure}
In this paper, we focus on the classical causal graph depicted in Fig.\ref{fig:graph_causal}, often used in the counterfactual fairness literature  \citep{kusner2017counterfactual,pfohl2019counterfactual,chiappa2019path}, which can apply for most applications. For more specific tasks, note further that our approach could be easily adapted for different graphs, such as those explored in \cite{kusner2017counterfactual} for instances. 
In this causal graph, both input $X$ and outcome $Y$ only depend on the sensitive attribute $A$ and a latent variable $U$, 
which represents all the relevant knowledge non dependent on the sensitive feature $A$. 
In that setting, the knowledge of $U$ can be used during training to simulate various versions of the same individual, corresponding to different values of $A$, in order to obtain a predictive function $h_\theta$ which respects the fairness objective from definition \ref{def1}. For any training individual, $U$ has to be inferred since only $X$, $A$ and $Y$ are observed. This inference  must however ensure that no dependence is created between $U$ and $A$ (no arrow from $U$ to $A$ in the graph from Fig.\ref{fig:graph_causal}), unless preventing the generation of proper alternative versions of $X$ and $Y$ for any values $A$. 

A classic way to achieve a counterfactually fair model is to proceed with two distinct main steps of Causal Inference and Model Learning  \citep{russell2017worlds,pfohl2019counterfactual},  that are described below. 

\subsubsection{Step 1: Counterfactual Inference}

The goal 
is to define a way to generate counterfactual versions of original individuals. As discussed above, this is usually done via approximate Bayesian inference, according to a pre-defined causal graph.

The initial idea  to  perform inference 
was to suppose with strong hypothesis a non deterministic structural model with some specific distribution for all the causal links  \citep{kusner2017counterfactual}. In this setting, the posterior distribution of $U$ was  estimated using the probabilistic programming language Stan \citep{team2016rstan}. Then, leveraging recent developments for approximate inference with deep learning, many works 
\citep{chiappa2019path,pfohl2019counterfactual,madras2019fairness,louizos2017causal} proposed to use Variational Autoencoding \cite{kingma2013auto} methods (VAE) to generalize this first model and capture more complex - non linear - dependencies in the causal graph. 

Following the  formulation of  VAE, it would be possible to directly optimize the classical lower bound (ELBO) \citep{kingma2013auto} on the training set $\cal D$, by minimizing:
\begin{align}
\mathcal{L}_{ELBO} = & -\mathbb{E}_{\substack{(x,y,a)\sim\mathcal{D}, \\ u \sim q_{\phi}(u|x,y,a)}}[\log p_{\theta}(x,y|u,a)]  \\   
& 
+ D_{KL}(q_{\phi}(u|x,y,a)||p(u))\big] \nonumber
\end{align}
where $D_{KL}$ denotes the Kullback-Leibler divergence of the posterior $q_{\phi}(u|x,y,a)$ from a prior $p(u)$, typically a standard Gaussian distribution ${\cal N}(0,I)$. The posterior  $q_{\phi}(u|x,y,a)$ is represented by a deep neural network with parameters $\phi$, which typically outputs the mean $\mu_\phi$ and the variance $\sigma_\phi$ of a diagonal Gaussian distribution ${\cal N}(\mu_\phi,\sigma_\phi I)$. The likelihood term factorizes as  
$p_{\theta}(x,y|u,a)=  p_{\theta}(x|u,a)p_{\theta}(y|u,a)$, which are defined as neural networks 
with parameters $\theta$. Since attracted by a  standard prior, the posterior is supposed to remove  probability mass for any features of the latent representation $U$ that are not involved in the reconstruction of $X$ and $Y$. Since $A$ is given together with $U$ as input of the likelihoods, all the information from $A$ should be removed from the posterior distribution of $U$.


However, many state of the art algorithms \citep{chiappa2019path,louizos2017causal,madras2019fairness,pfohl2019counterfactual} show that the independence level between the latent space $U$ and the sensitive variable $A$ is insufficient with this classical ELBO optimization. Some information from $A$ leaks in the inferred $U$. In order to ensure a high level of independence, a specific TARNet \citep{shalit2017estimating} architecture can be employed  \citep{madras2019fairness} or a penalisation term can be added in the loss function. For example, \citep{chiappa2019path,pfohl2019counterfactual} add a Maximum Mean Discrepancy (MMD) \citep{gretton2012kernel} constraint. The MMD term can be used to enforce all the different aggregated posterior to the prior distribution\citep{pfohl2019counterfactual}: $\mathcal{L}_{MMD}(q_{\phi}(u|A=a_{k})||p(u))$ for all $a_{k} \in \Omega_{A}$ (referred to as MMD wrt $P(U)$ in the following). 
Alternatively, the constraint can directly enforce the matching between pairs of posteriors \citep{chiappa2019path}:
$\mathcal{L}_{MMD}(q_{\phi}(u|A=a_{k})||q_{\phi}(u|A=a))$ for all $a_{k} \in \Omega_{A}$, with $a$ standing for the original sensitive value of the considered individual (referred to as MMD wrt $U_a$ in the following).
Notice that while 
this additional term can improve independence, it can also encourage the model to ignore the latent confounders $U$, by being too restrictive. 
One possible approach to address this issue is to apply weights $\lambda$ 
(hyperparameters) to control the relative importance of the different terms. In addition, we employ in this paper a variant of the ELBO optimization as done in  \citep{pfohl2019counterfactual}, where the $D_{KL}(q_{\phi}(u|x,y,a)||p(u))$ term is replaced by a MMD term $\mathcal{L}_{MMD}(q_{\phi}(u)||p(u))$ between the aggregated posterior $q_{\phi}(u)$ and the prior. This has been shown more powerful than the classical $D_{KL}$  
for ELBO optimization in \citep{zhao2017infovae}, as the latter can reveal as 
too restrictive (uninformative latent code problem) \citep{chen2016variational,bowman2015generating,sonderby2016ladder} and can also tend to overfit the data (Variance Over-estimation in Feature Space). 
Finally, the inference for counterfactual fairness can be optimized by minimizing 
\citep{pfohl2019counterfactual}:
\begin{align}
\mathcal{L}_{CE-VAE} = & - \mathop{\mathbb{E}}_{\substack{(x,y,a)\sim\mathcal{D}, \\ u \sim q_{\phi}(u|x,y,a)}}\left[\begin{array}{l}  \lambda_x \log(p_{\theta}(x|u,a)) \ +
\nonumber \\ 
 \lambda_y \log(p_{\theta}(y|u,a))\end{array} \right]  \nonumber \\ +&\lambda_{MMD} \  \mathcal{L}_{MMD}(q_{\phi}(u)||p(u))
\label{overallLoss3}
 \\+&\lambda_{ADV} \ \frac{1}{m_{a}}\sum_{a_{k} \in \Omega_{A}} \mathcal{L}_{MMD}(q_{\phi}(u|a=a_{k})||p(u))\nonumber
\end{align}
where $\lambda_{x}$, $\lambda_{y}$, $\lambda_{MMD}$, $\lambda_{ADV}$ are scalar hyperparameters and $m_{a}=|\Omega_{A}|$. The additional MMD objective  
can be interpreted as minimizing the distance between all moments of each aggregated latent code distribution and the prior distribution, in order to remove most sensitive dependency 
from the code generator. It requires however a careful design of the kernel used for MMD computations (typically a zero mean isotropic Gaussian).  Note that we chose to present all models with a generic inference scheme $q(U|X,Y,A)$, while most approaches from the literature only consider $q(U|X,A)$. The use of $Y$ as input is allowed since $U$ is only used during training, for generating  counterfactual samples used to learn the predictive model in step 2. Various schemes of inference are considered in our experiments (section \ref{xp}).









\subsubsection{Step 2: Counterfactual predictive model} Once the causal model is learned, the goal is to use it to learn a fair predictive function $h_\theta$, by leveraging the ability of the model to generate alternative versions of each training individual. 
The global loss function is usually composed of the traditional predictor loss $l(h_{\theta}(x_{i},a_{i}),y_{i})$ (e.g. cross-entropy for instance $i$) and the counterfactual unfairness estimation term $\mathcal{L_{CF}(\theta)}$:  
\begin{equation}
 \mathcal{L} =
 \frac{1}{m} \sum_i^{m} l(h_{\theta}(x_i),y_i)
 +  \lambda \mathcal{L_{CF}(\theta)}
 \label{loss:lossfunc}
\end{equation}
where  $\lambda$ is an  hyperparameter which controls the impact of the counterfactual loss in the optimization. 
The counterfactual loss $\mathcal{L_{CF}(\theta)}$ considers differences of predictions for alternative versions of any individual.   
For example, 
\citep{russell2017worlds} 
considers the following Monte-Carlo estimate from 
$S$ samples for each individual $i$ and each value $a \in \Omega_{A}$:
\begin{equation}
\label{lcf}
\mathcal{L_{CF}(\theta)}= \frac{1}{m}\sum_{i=1}^{m}\frac{1}{m_{a}}\sum_{a_{k} \in \Omega_{A}}\frac{1}{S}\sum_{s=1}^{S} \Delta_{a_k}^{i,s}
\end{equation}
\normalsize
where $\Delta_{a_k}^{i,s}=\Delta(h_{\theta}(x_{i,A\leftarrow{a_i}}^{s},a_i),h_{\theta}(x_{i,A\leftarrow{a_{k}}}^{s},a_k))$ is a loss function that compares two predictions, $x_{i,A\leftarrow{a}}^{s}$ denotes the s-th sample from the causal model for the i-th individual of the training set and the sensitive attribute value $a$. Following the causal model learned at step 1, $x_{i,A\leftarrow{a}}^{s}$ is obtained by first inferring a sample $u$ from $q_{\phi}(u|x_i,a_i,y_i)$ and then sampling $x_{i,A\leftarrow{a}}^{s}$ using $p_{\theta}(x|u,a)$ with the counterfactual (or factual) attribute value $a$.  According to the task, $\Delta$ can take various forms. For binary classification, it can correspond to a logit paring loss as done in \cite{pfohl2019counterfactual}:  $\Delta(z,z')=(\sigma^{-1}(z)-\sigma^{-1}(z'))^2$, where $\sigma^{-1}$ is the logit function. For continuous outcomes, it can simply correspond to a mean squared difference. 
\subsubsection{Discussion}
For now, state-of-the-art approaches have focused  specifically on categorical variables $A$. 
Unfortunately, the classical methodology for CounterFactual Fairness as described above  cannot be directly generalized 
for continuous sensitive attributes,  
because the two steps involve enumerations of the discrete counterfactual modalities $a_{k}$ in the set $\Omega_{A}$. 
Particularly in step 1, sampling $A$ from a uniform distribution for approximating the expectation $E_{a \sim p(A)}\mathcal{L}_{MMD}(q_{\phi}(u|A=a)||p(u))$ is not an option since this requires to own a good estimation of $q_{\phi}(u|A=a)$ for any $a \in \Omega_A$, which is difficult in the continuous case. While such a posterior  can be obtained for discrete sensitive attributes (at least when $|\Omega_A| << m$) by aggregating the posteriors $q_{\phi}(u|x_i,a_i,y_i)$ over training samples $i$ such that $a_i=a$, such a simple aggregation over filtered samples is not possible for continuous attributes.  Moreover, existing approaches based on MMD costs imply to infer codes $U$ from a distribution that takes $A$ as input, in order to be able to obtain the required aggregated distributions via:
$q_\phi(u|a) = \mathbb{E}_{p_{data}(x,y|a)} [q_\phi(u\vert x,y,a)]$. Omitting $A$ from the conditioning of the generator would correspond to assume the mutual independence of $u$ and $a$ given $x$ and $y$, which is usually wrong. On the other hand, passing $A$ to the generator of $U$ can encourage their mutual dependency in some settings, as we observe in our experiments.  






\section{Adversarial learning for counterfactual fairness} 
\label{model}


In this section we revisit the 2 steps shown above by using adversarial learning rather than MMD costs for ensuring Counterfactual Fairness. Our contribution covers a broad range of scenarios, where the sensitive attribute $A$ and the outcome value $Y$ can be either discrete or continuous.



\subsubsection{Step 1: Counterfactual Inference}

\begin{figure}[h]
  \centering
  \includegraphics[scale=0.45,valign=t]{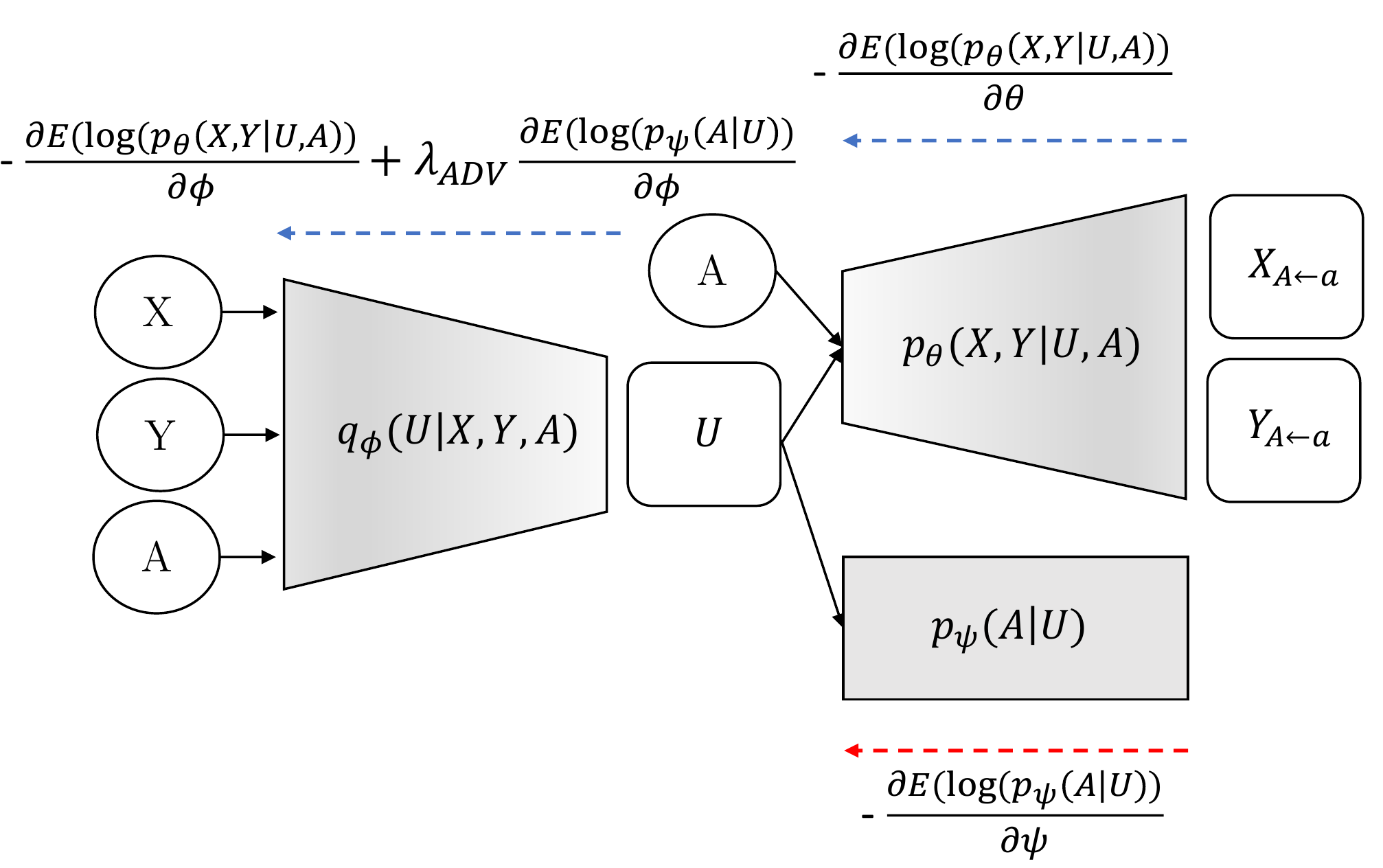}
  \caption{Architecture of our Counterfactual inference process. 
  Blue arrows represent the retro-propagated gradients for the minimization of the global objective.  The red one corresponds to the gradients for the adversarial optimization. Circles are observed variables, squares are samples from the neural distributions.}
  \label{fig:Adversarial_fig}
\end{figure}

To avoid the comparison of distributions for each possible sensitive value, which reveals particularly problematic in the continuous setting, we propose to employ an adversarial learning framework, which allows one to avoid the enumeration of possible values in $\Omega_A$. We follow an  approach similar to the adversarial auto-encoders proposed in \cite{MakhzaniSJG15}, but where the discriminator real/fake data is replaced by a sensitive value predictor.  
The idea is to avoid any adversarial function to be able to decode $A$ from the code $U$ inferred from the encoder $q_\phi$, which allows one to ensure mutual independence of $A$ and $U$.   
This defines a two-players adversarial game, such as in GANs \citep{Goodfellow2014}, where the goal is to find some parameters $\phi$ which minimize the loss to reconstruct $X$ and $Y$, while maximizing the reconstruction loss of $A$ according to the best decoder $p_{\psi}(A|U)$: 
\begin{equation}
\label{ourInf}
\centering
\argmin_{{\theta,\phi}}\max_{{\psi}} \mathcal{L}_{ADV}(\theta,\phi,\psi)
\end{equation}

with, for the graphical causal model from figure \ref{fig:graph_causal}: 

\begin{align}
\mathcal{L}_{ADV}(\theta,\phi,\psi) =& - \mathop{\mathbb{E}}_{\substack{(x,y,a)\sim\mathcal{D}, \\ u \sim q_{\phi}(u|x,y,a)}}\left[\begin{array}{l}  \lambda_x \log(p_{\theta}(x|u,a)) \ +
\nonumber \\ 
 \lambda_y \log(p_{\theta}(y|u,a))\end{array} \right]  \nonumber \\  +& \lambda_{MMD} \  \mathcal{L}_{MMD}(q_{\phi}(u)||p(u))  \\  +& \lambda_{ADV} \nonumber \  \mathop{\mathbb{E}}_{\substack{(x,a)\sim\mathcal{D}, \\ u \sim q_{\phi}(u|x,y,a))}}[log(p_{\psi}(a|u))]
\end{align}
where $\lambda_{x}$, $\lambda_{y}$, $\lambda_{MMD}$, $\lambda_{ADV}$ are scalar hyperparameters. Compared to existing approaches presented in previous section, the difference is the last term which corresponds to the expectation of the log-likelihood of $A$ given $U$ according to the decoder with parameters $\phi$. This decoder corresponds to a neural network which outputs the parameters of the distribution of $A$ given $U$ (i.e., the logits of a Categorical distribution for the discrete case, the mean and log-variance of an diagonal Gaussian in the continuous case).


All parameters are learned conjointly. Figure \ref{fig:Adversarial_fig} gives the full architecture of our variational adversarial inference for the 
causal model from figure \ref{fig:graph_causal}. It depicts the neural network encoder $q_{\phi}(U\lvert X,Y,A)$ 
which generates a latent code $U$ from the inputs $X$, $Y$ and $A$. A neural network decoder $p_{\theta}(X,Y\lvert U)$ reconstructs the original $X$ and $Y$ from both $U$ and $A$. The adversarial  network $p_\psi$ 
tries to reconstruct the sensitive attribute $A$ from the confounder $U$. 
As classically done in adversarial learning, we alternate steps for the adversarial maximization and steps of global loss minimization (one gradient descent iteration on the same batch of data at each step). 
Optimization is done via the re-parametrization trick \cite{kingma2013auto} to handle stochastic optimization. 

\subsubsection{Step 2: Counterfactual predictive model}
As described in section 2.3,  the counterfactual fairness in the predictive model learned at step 2 is ensured by comparing, for each training individual,  counterfactual predictions $Y_{A \leftarrow a'}$ for all $a' \in \Omega_A$. For the discrete case (i.e., $A$ is a Categorical variable), we keep this process for our experiments. However, for the continuous setting (i.e., $A$ is for instance generated from a Gaussian), such an approach must be somehow adapted, due to the infinite set $\Omega_{A}$. In that case, we can consider a sampling distribution $P'(A)$ to formulate the 
following loss, which can be optimized via Monte-Carlo sampling and stochastic gradient descent (SGD):  
\begin{multline}
\label{lcf2}
\mathcal{L}_{CF}(\theta) =  \frac{1}{m} \sum_i^{m}  l(h_{\theta}(x_i),y_i)  \\ + \lambda \mathop{\mathbb{E}}_{\substack{u \sim P(u|x_i,a_i,y_i), \\ \tilde{x} \sim P(x|u_i,a_i), \\   a'\sim P'(A),  x' \sim P(x|u,a')}}[(h_{\theta}(\tilde{x})  - h_{\theta}(x'))^2] 
\end{multline}

This formulation is equivalent to the one from Eq. \ref{lcf}, for continuous outcomes $\hat{Y}$ (thus considering a least squared cost as $\Delta$)  
and for continuous attributes $A$ (thus using the sampling distribution $P'(A)$ rather than considering every possible $a \in \Omega_A$). 

Note that using a non-uniform sampling distribution 
$P'(A)$ would enforce the attention of the penalisation near the mass of the distribution. This prevents using the prior of $A$ estimated from the training set, since this would tend to reproduce inequity between individuals: counterfactual predictions for rare $A$ values would be be little taken into account during training. We therefore consider a uniform $P'(A)$ in our experiments for the continuous setting when using the $ \mathcal{L}_{CF}(\theta)$ objective at step 2.



However, for the specific case of 
high-dimensional sensitive attributes $A$, using a uniform sampling distribution $P'(A)$ could reveal as particularly inefficient. The risk is that a high number of counterfactual samples fall in easy areas for the learning process, while some difficult  areas - where an important  work for fairness has to be performed - remain insufficiently visited.

%
To tackle this problem, we propose to allow the learning process to dynamically focus on the most useful areas of $\Omega_A$ for each individual. During learning, we consider an adversarial process, which is in charge of moving the sampling distribution $P'(A)$, so that the counterfactual loss is the highest. This allows the learning process to select useful counterfactuals for ensuring fairness. Who can do more can do less: dynamically focusing on hardest areas allows one to expect fairness everywhere. Again, we face a  two-players adversarial game, which formulates as follows:  
\begin{equation}
\label{dyn}
\centering
\argmin_{{\theta}}\argmax_{{\phi}}\mathcal{L}_{DynCF}(\theta,\phi) 
\end{equation}
with:
\begin{align}
\label{dyn2}
    \mathcal{L}_{DynCF}(\theta,\phi) = & \frac{1}{m} \sum_i^{m}  l(h_{\theta}(x_i),y_i)  \\  & + \lambda \hspace*{-0.2cm} \mathop{\mathbb{E}}_{\substack{u \sim P(u|x_i,a_i,y_i), \\  \tilde{x} \sim P(x|u,a_i), \\ a' \sim P_\phi(a|u),    x' \sim P(x|u,a')}} \hspace*{-0.3cm} [(h_{\theta}(\tilde{x})  -  h_{\theta}(x'))^2] \nonumber
\end{align}

Compared to Eq. \ref{lcf2}, this formulation considers an adversarial sampling distribution $P_\phi(A|U)$ rather than a uniform static distribution $P'(A)$. It takes the form of a neural network that outputs the parameters of the sampling distribution for a given individual representation $U$. In our experiments we use a diagonal logit-Normal 
distribution  $sigmoid(\mathcal{N}(\mu_{\phi}(u),\sigma_{\phi}^2(u)I))$, where $\mu_{\phi}(u)$ and $\sigma_{\phi}^2(u)$ stand for the mean and variance parameters provided by the network for the latent code $u$. Samples from this distribution are then projected on the support $\Omega_A$ via a linear mapping depending on the shape of the set. Passing $U$ as input for the network allows the process to define different distributions for different codes: according to the individual profiles, the unfair areas are not always the same. This also limits the risk that the adversarial process gets stuck in sub-optimums of the sensitive manifold. As done for adversarial learning in step 1, all parameters are learned conjointly, by alternating steps for the adversarial maximization and steps of global loss minimization. 
The re-parametrization trick \cite{kingma2013auto} is also used, for the adversarial optimization of $P_\phi(A|U)$. 

\section{Experiments}
\label{xp}

We empirically evaluate the performance of our contribution on  6 real world data sets. 
For the discrete scenario and specifically in the binary case ($Y \in \{0,1\}, A \in \{0,1\}$), we use 3 different popular data sets: the Adult UCI income data set~\citep{Dua:2019} with a gender sensitive attribute (male or female), the COMPAS data set~\citep{angwin2016machine}  with the race sensitive attribute (Caucasian or not-Caucasian) and the Bank dataset~\citep{Moro2014} with the age as sensitive attribute (age is between 30 and 60 years, or not). For the continuous setting ($Y$ and $A$ are continuous), we use the 3 following data sets: the US Census dataset \citep{USCensus} with gender rate as sensitive attribute encoded as the percentage of women in the census tract, the Motor dataset \citep{pricinggame15} with the driver’s age as sensitive attribute and the Crime dataset \citep{Dua:2019} with the ratio of an ethnic group per population as sensitive attribute. 
\normalsize

Additionally to the 6 real-world datasets, we consider a synthetic scenario, that allows us to perform a further analysis of the relative performances of the approaches.  
The synthetic scenario subject is a pricing algorithm for a fictional car insurance policy, which follows the causal graph from figure \ref{fig:graph_causal}. We simulate both a binary and a continuous dataset from this  scenario. The main advantage of these synthetic scenarios is that it is possible to get "ground truth" counterfactuals for each code $U$, obtained using the true relationships of the generation model while varying A uniformly in $\Omega_A$. This will allow us to evaluate the counterfactual fairness of the models without depending on a given inference process for the evaluation metric, by relying on prediction differences between these true counterfactuals and the original individual.
The objective of this scenario is to achieve a counterfactual fair predictor which estimates the average cost history of insurance customers. We suppose 5 unobserved variables (Aggressiveness, Inattention, Restlessness, Reckless and Overreaction) which corresponds to a 5 dimensional confounder $U$. The input X is composed of four explicit variables $X_1,...,X_4$ which stand for  
vehicle age, speed average, horsepower and average kilometers per year respectively. We consider the policyholder's age as sensitive attribute $A$. The input $X$ and the average cost variable $Y$ are sampled from $U$ and $A$ as depicted in figure 1 from the main paper. We propose both a binary and a continuous version of this  scenario. For both of them, 5000 individuals are sampled. 
Details of distributions used
for the continuous setting of this synthetic scenario are given below: 
 {
\begin{align*}
\centering
U &\sim \mathcal{N}
\begin{bmatrix}
\begin{pmatrix}
0\\
0.5\\
1\\
1.5\\
2
\end{pmatrix}\!\!,&
\begin{pmatrix}
1 & 0 & 0 & 0 & 0 \\
0 & 4 & 0 & 0 & 0 \\
0 & 0 & 2 & 0 & 0 \\
0 & 0 & 0 & 3 & 0 \\
0 & 0 & 0 & 0 & 2 
\end{pmatrix}
\end{bmatrix}\\[2\jot]
X1 &\sim \mathcal{N}(7+0.1*A+ U_{1}+U_{2}+U_{3},1) ;\\ 
X2 &\sim \mathcal{N}(80+ A + U_{2}^{2},10) ; \\
X3 &\sim \mathcal{N}(200+5*A + 5*U_{3},20) ; \\
X4 &\sim \mathcal{N}((10^4+5*A + U_{4}+U_{5},1000) \\
X &\sim [X1,X2,X3,X4]; \\
A &\sim \mathcal{N}[45,5]; \\
Y &\sim {\cal N}( 2*(7*A + 20*\sum_j U_{j}),0.1)
\end{align*}}

\subsection{Step 1: Counterfactual Inference} 
In this section, we report experiments performed for assessing our adversarial approach for  Counterfactual Inference (step 1 of the previous section). 
We compare our adversarial approach with two version of the approach in Eq.~\ref{overallLoss3}, each using one of the two MMD constraints  MMD wrt $P(A)$ or MMD wrt $U_a$ as presented in section \ref{counterfact} (step 1). Note that these approaches are not applicable for continuous datasets as discussed at the end of section \ref{bg}.  
For every approach, we compare three different inference schemes for $U$: $q_{\phi}(u|x,y,a)$, $q_{\phi}(u|x,y)$ and $q_{\phi}(u|x,a)$. 
As a baseline, we also use a classical Variational Autoencoder inference without counterfactual independence constraint 
(i.e., Eq. \ref{counterfact} without the last term). 

All hyper-parameters for every approach have been tuned by  5-fold cross-validation. For the US Census data set for our approach for instance, the encoder $q_\phi$ architecture is an MLP of 3 hidden layers with 128, 64 and 32 units respectively, with ReLU activations. 
On this dataset, the decoder $p_\theta$ is an MLP of only  one hidden layer with 64 units with a ReLu activation function and the output consists in one single output node with linear activation to reconstruct $Y$ and $37$ units to reconstruct $X$ (number of features). The adversarial neural network $p_\psi$ is an MLP of two hidden layers with 32 and 16 units respectively. 
For the binary datasets, a sigmoid is applied on the outputs of decoders for $A$ and $Y$.  For both MMD constraints we used a Gaussian radial basis function kernel. 
For all datasets, the prior distribution $p(U)$ considered for training the models is a  five-dimensional standard Gaussian.

In order to evaluate the level of dependence between the latent space $U$ and the sensitive variable $A$, we compare the different approaches by using the neural estimation of the $HGR$ correlation coefficient given in \citep{abs-1911-04929}. This coefficient, as shown above in Eq.\ref{hgr}, assesses the level of non-linear dependency between two jointly distributed random variables. The estimator is trained for each dataset and each approach on the train set, comparing observed variables $A$ with the corresponding inferred codes $U$.  

For  all  data sets,  we  repeat  five  experiments  by  randomly sampling two subsets, $80\%$ for the training set and $20\%$ for the test set. Finally, we report the average reconstruction loss for $X$ and $Y$ on the test set, as long as the HGR between inferred test codes and the corresponding sensitive attributes.  Results of our experiments can be found in table \ref{tab:binary_case} for the discrete  case and table \ref{tab:continuous_case} for the continuous case. For all of them, we attempted via the different hyperparameters ($\lambda_{x}$, $\lambda_{y}$, $\lambda_{MMD}$, $\lambda_{ADV}$) to obtain the lower dependence measure while keeping the minimum loss as possible to reconstruct $X$ and $Y$.

\begin{figure*}[h!]
\centering
\subfloat[\small $\lambda=0.00$ ; $CF=72\%$]{\label{fig:mdleft}{\includegraphics[scale=0.396,valign=t]{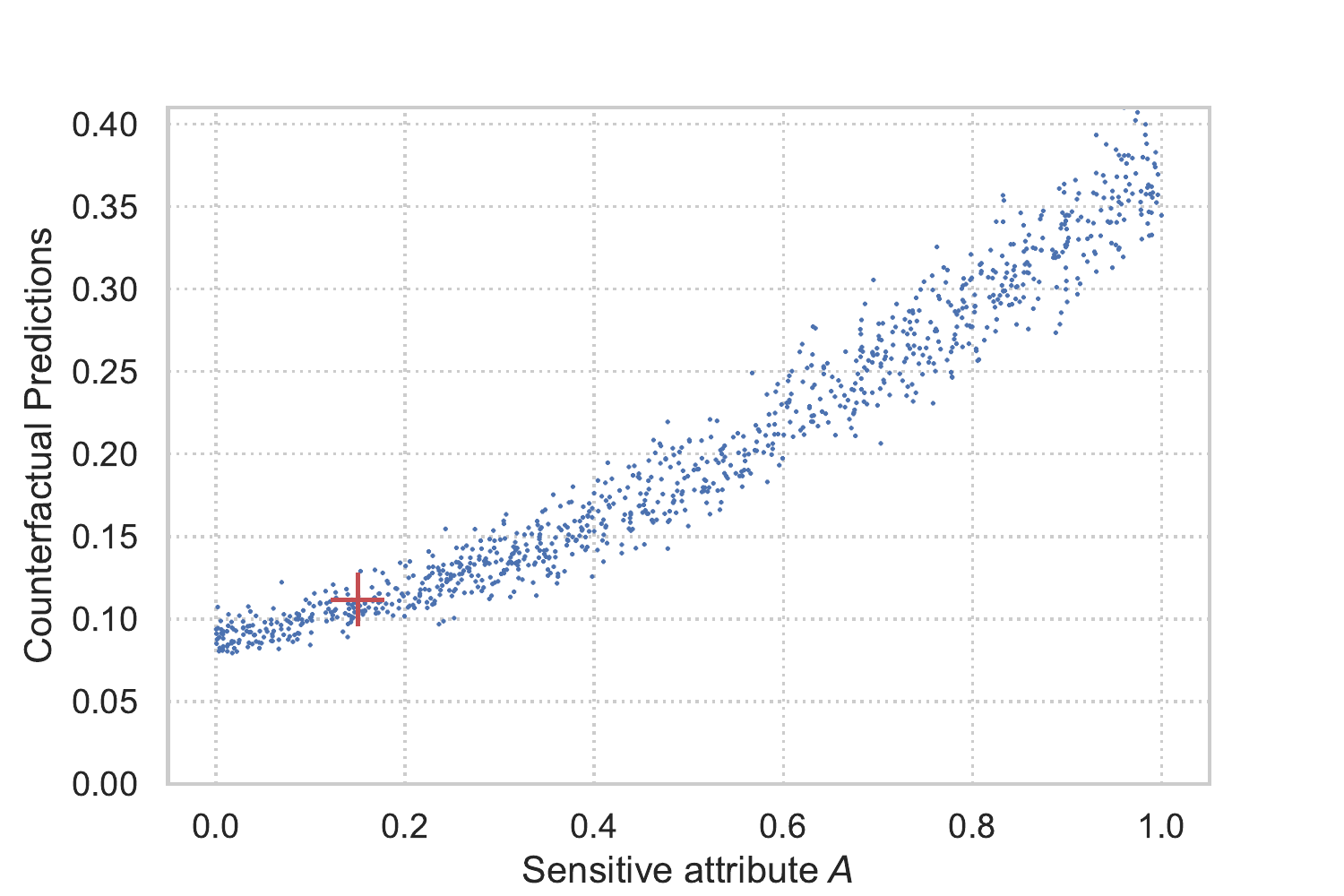}}}\hfill
\subfloat[\small $\lambda=0.10$ ; $CF=44\%$]{\label{fig:mdmidle}{\includegraphics[scale=0.396,valign=t]{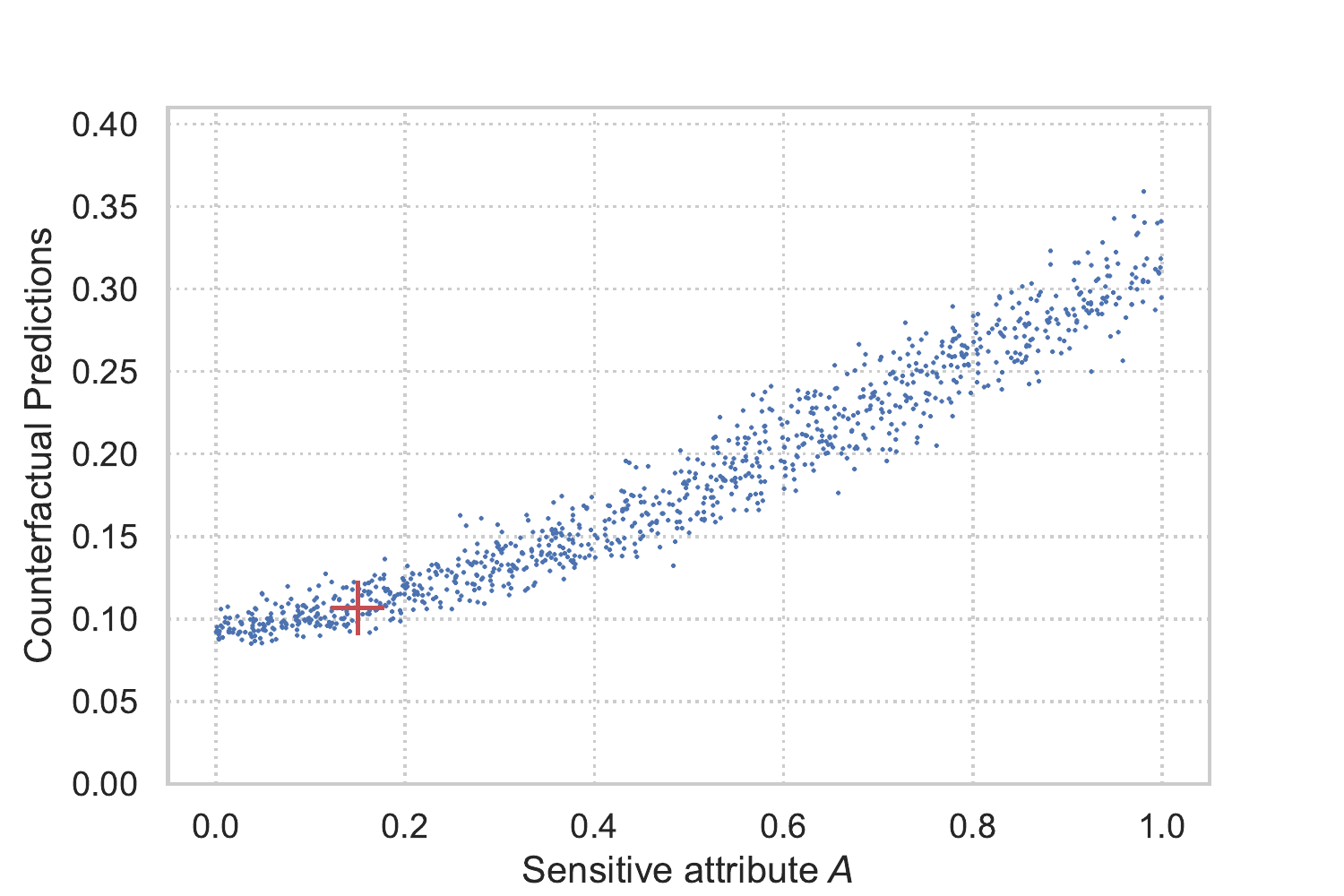}}}\hfill
\subfloat[\small $\lambda=0.30$ ; $CF=04\%$]{\label{fig:mdright}{\includegraphics[scale=0.396,valign=t]{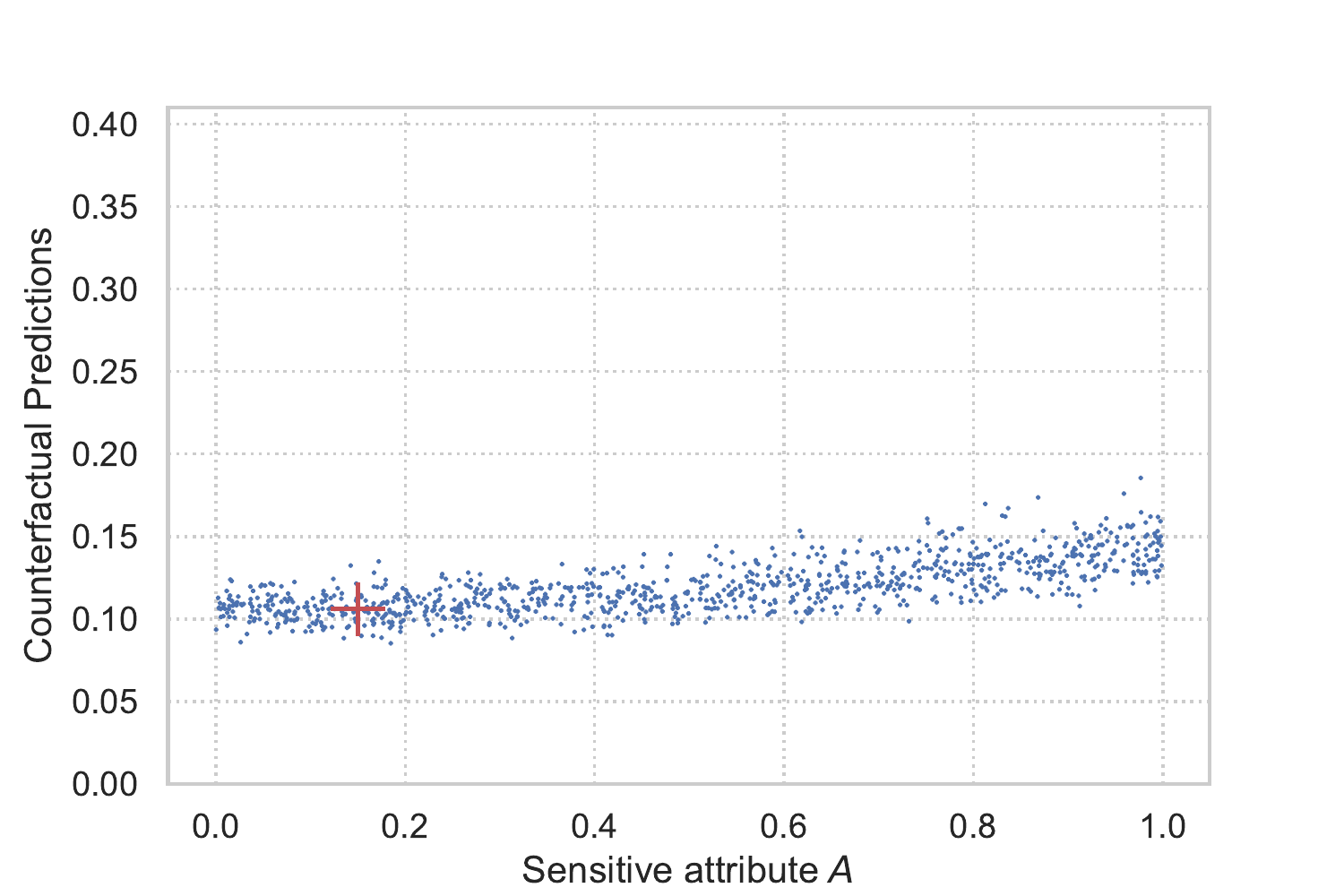}}}
\caption{Impact of $\lambda$ (Crime data set) on a specific instance $i$. Blue points are  counterfactual predictions $h_{\theta}(x_{i,A\leftarrow{a'}})$ from $1.000$ points $A\leftarrow{a'}$  generated randomly. The red cross represents the prediction $h_{\theta}(x_{i,A\leftarrow{a}})$ for the real $A=a$ of  instance $i$. }
\label{fig:Error_wrt_A}
\end{figure*}

\begin{table*}[h!]
\caption{Inference results in the discrete case}
\label{tab:binary_case}
\centering
\resizebox{\textwidth}{!}{
\begin{tabular}{||l|l||l|l|l||l|l|l||l|l|l||l|l|l|l||}
\hhline{==============} 
 \multicolumn{2}{||c||}{\multirow{2}{*}{ }} & \multicolumn{3}{c||}{Adult UCI}      & \multicolumn{3}{c||}{Compas}                                &   \multicolumn{3}{c||}{Bank}    
   &   \multicolumn{3}{c||}{Synthetic Scenario}     
   \\ \cline{3-14}
   \multicolumn{2}{||c||}{ } & Loss X & Loss Y & HGR
&  Loss X & Loss Y & HGR
& Loss X & Loss Y   & HGR 
& Loss X & Loss Y   & HGR 
\\ \hline
\multirow{4}{*}{\rotatebox[origin=c]{90}{($\{x,y,a\}$)}}
&  No Constraint, $q(u|x,y,a)$
&  0.0781  &   0.0006  &  0.6984 
&   0.0278  &   0.0041 & 0.6952  
&  0.0963  &  0.0001 & 0.5988  
&  0.2681  &  0.0085  &  0.9725
\\
&  Adv. Constraint, $q(u|x,y,a)$ &   0.1091  &  0.0009 &  \textbf{0.5453}  
&  0.0254  &   0.0020  &  \textbf{0.2693}   
&  0.2038  &  0.0005 &   \textbf{0.3423}   
&  0.2669  &  0.0721  &  \textbf{0.4167}
\\
& MMD wrt $P(U)$, $q(u|x,y,a)$
& 0.1286   &   0.0012   & 0.7017  
&   0.0252     & 0.0029     &  0.6565  
&  0.2002  & 0.0002 &   0.4521  
&  0.2535  &  0.0839  &   0.6623
\\ 
& MMD wrt $U_{a}$, $q(u|x,y,a)$
& 0.0938   &      0.0009  &  0.7181  
& 0.0259   &   0.0098   &  0.8892    
&  0.1263  &  0.0003  &   0.5188   
& 0.2762  &  0.0351  &  0.5697 
\\ \hline
\multirow{4}{*}{\rotatebox[origin=c]{90}{($\{x,y\}$)}}
& No Constraint, $q(u|x,y)$  
&  0.0786  &   0.0008   &  0.6077 
&  0.0274     & 0.0133   &  0.3817 
&   0.0957 &  0.0001 &  0.4989 
&  0.2577  &  0.0022 & 0.6418
\\ 
& Adv. Constraint, $q(u|x,y)$
&   0.1272  &  0.0329    &  \textbf{0.1811} 
&   0.0245  &  0.0013  &  \textbf{0.1728}   
& 0.1858   &  0.0073 &     \textbf{0.2476} 
&  0.2649  &  0.1015 &  0.4521
\\ & MMD wrt $P(U)$, $q(u|x,y)$ 
&   0.1287  &  0.0016    &  0.6092  
&  0.0259    &  0.0055   &  0.4470  
&  0.1898  & 0.0003 &   0.3716  
&  0.2567  & 0.0885  & 0.6868
\\ & MMD wrt $U_{a}$, $q(u|x,y)$
&   0.0872  &    0.0013  &  0.6852  
&   0.0266   &   0.0094  &  0.3109  
&   0.1415  &  0.0003  &   0.3929  
& 0.2674  & 0.0553  & \textbf{0.4473}
\\ \hline
\multirow{4}{*}{\rotatebox[origin=c]{90}{($\{x,a\}$)}}
& No Constraint, $q(u|x,a)$
&  0.0982  &   0.3534  &   0.6689  
&  0.0288    &  0.8246  &  0.3726 
&  0.1391  & 0.2101  &  0.5572  
&  0.2686 &  0.0128 & 0.7040
\\ & Adv. Constraint, $q(u|x,a)$   
&    0.0995   &   0.3462   & 0.5259   
&   0.0271    &  0.6889   &  0.4344    
& 0.1880   &  0.2110  &  \textbf{0.3061}     
&  0.2589  &  0.0980 & \textbf{0.4264}
\\ & MMD wrt $P(U)$, $q(u|x,a)$
&   0.1308  &  0.3559    &   \textbf{0.3586}  
&   0.0288  &  0.7611   & 0.4365  
&   0.2141   &  0.2129  &    0.3386 
&  0.2506  & 0.1176  & 0.6298
\\ & MMD wrt $U_{a}$, $q(u|x,a)$   
&  0.0940  &   0.3603   &  0.5811  
&   0.0278 &  0.7314  &  \textbf{0.3345}    
&   0.1485  &  0.2135  &   0.5536 
&  0.2584  & 0.1076 &  0.4692
\\
\hhline{===============} 
\end{tabular}
}
\\
\end{table*}
\begin{table*}[h!]
\caption{Inference results in the continuous case}
\label{tab:continuous_case}
\centering
\resizebox{\textwidth}{!}{
\begin{tabular}{l|l|l|l|l|l|l|l|l|l|l|l|l|l}
\cline{2-13}

& \multicolumn{3}{c|}{US Census}                     & \multicolumn{3}{c|}{Motor}                                & 
\multicolumn{3}{c|}{Crime}                                      & 
\multicolumn{3}{c|}{Synthetic Scenario} 
\\ \cline{2-13} 
& \multicolumn{1}{c|}{Loss X} & \multicolumn{1}{c|}{Loss Y} & \multicolumn{1}{c|}{HGR} &   \multicolumn{1}{c|}{Loss X} & \multicolumn{1}{c|}{Loss Y} & \multicolumn{1}{c|}{HGR}& \multicolumn{1}{c|}{Loss X} & \multicolumn{1}{c|}{Loss Y}   & \multicolumn{1}{c|}{HGR} & \multicolumn{1}{c|}{Loss X} & \multicolumn{1}{c|}{Loss Y}   & \multicolumn{1}{c|}{HGR}
\\ \hline
\multicolumn{1}{|l|}{No Cons. $q(u|x,y,a)$}   &  0.1685 &  0.0019  &   0.5709  &  0.2526   &  0.0024    &   0.9023   &  0.4558  &   0.0016 &   0.9059    & 0.6788  & 0.0076 &   0.9523   \\ \hline
\multicolumn{1}{|l|}{No Cons. $q(u|x,y)$}    &   0.1690    &  0.0005  &   0.4163   &   0.3068  &   0.0034  &  0.9479 & 0.4523  &  0.0018  &  0.8998  & 0.6495 & 0.0003  &   0.6227 \\ \hline
\multicolumn{1}{|l|}{No Cons.  $q(u|x,a)$}     &  0.1726  &  0.2886  &  0.8252  &    0.3377  &  0.9381  &  0.9728   &  0.4634   &   0.3999    & 0.9076  & 0.6751 & 0.4554 &  0.8650 \\ \hline
\multicolumn{1}{|l|}{Adv $q(u|x,y,a)$}   &  0.1617  &  0.0004 & 0.3079 &  0.4702  &  0.0035   &   0.2941  &  0.4865   &   0.0701    &  0.5268    & 0.6804 & 0.0088 & 0.2280  \\ \hline
\multicolumn{1}{|l|}{Adv $q(u|x,y)$}   &  0.1663  &  0.0009   & 0.2980    &  0.3694  &  0.0057  &  0.3314  &  0.4835  &    0.0571  &  0.6024 & 0.6633 & 0.1196 &  0.3175 \\ \hline
\multicolumn{1}{|l|}{Adv $q(u|x,a)$}   & 0.1828  & 0.2891  & 0.3285 &  0.4706  & 0.9878  &  0.2478  &   0.4904  &   0.3933  &  0.5810  & 0.6862  &  0.8819 & 0.5148  \\ \hline
\end{tabular}
}
\\ 
\end{table*}

As expected, the baseline without the independence constraint achieves the best $X$ and $Y$ reconstruction loss, but this is also the most biased one with the worst dependence in term of HGR in most datasets. Comparing the different constraints in the discrete case, the adversarial achieves globally the best result with the lower HGR while maintaining a reasonable reconstruction for $X$ and $Y$. It is unclear which MMD constraint performs better than the other. We observe that the best results in terms of independence are obtained without the sensitive variable  given as input of the inference network (inference only with $X$ and $Y$). Note however that for the MMD constraints, this setting implies to make the wrong assumption of independence of $U$ w.r.t. $A$ given $X$ and $Y$ for the estimation of the constraint (as discussed at the end of section \ref{bg}). This is not the case for our adversarial approach, which obtains particularly good results on this setting for discrete datasets. On continuous datasets, our approach succeeds in maintaining reasonable reconstruction losses for important gains in term of HGR compared to the classical VAE approach (without constraint). Interestingly, on these datasets, it appears that our approach obtains slightly better results when using the full information ($X$, $Y$ and $A$) as input of the inference network. We explain this by the fact that removing 
the influence of a binary input is harder than the one of a smoother continuous one, while this can reveal as a useful information for generating relevant codes. 



\subsection{Step 2: Counterfactual predictive model} This section reports experiments involving the training procedure from step 2 as described in section \ref{model}. The goal of these experiments is threefold: 1. assess the impact of the adversarial inference on the target task of counterfactual fairness, 2. compare our two proposals for counterfactual bias mitigation (i.e., using a uniform  distribution or an adversarial dynamic one for the sampling of counterfactual sensitive values) and 3. assess the impact of the control parameter from Eq.\ref{dyn}. 

 The predictive model used in our experiments is a MLP with 3 hidden layers. The adversarial network $P_\phi$ from Eq.\ref{dyn2} is a MLP with 2 hidden layers and RELU activation. For all our experiments, a single counterfactual for each individual is sampled at each iteration during the training of the models. 
 Optimization is performed using ADAM. 

Tables \ref{tab:binary_case2} and \ref{tab:continuous_case2} report results for the discrete and the continuous case respectively. The inference column refers to the inference process that was used for sampling counterfactuals for learning the predictive model. For each setting, we use the best configuration from tables \ref{tab:binary_case} and \ref{tab:continuous_case}. The mitigation column refers to the type of counterfactual mitigation that is used for the results: No mitigation or $L_{CF}$ (Eq.\ref{lcf}) for the discrete case; No mitigation, $L_{CF}$ (Eq.\ref{lcf2}) or $L_{DynCF}$ (Eq.\ref{dyn2}) for the continuous setting.   
Results are reported in terms of accuracy (for the discrete case) or MSE (for the continuous case) and of Counterfactual Fairness (CF). 
The CF measure 
is defined, for the $m_{test}$ individuals from the test set, as:  
\begin{equation}
\label{cf}
 CF=\frac{1}{m_{test}} \sum_i^{m_{test}} \mathbb{E}_{(x',a') \sim C(i)} [\Delta(h_{\theta}(x_i, a_i),h_{\theta}(x',a'))]
\end{equation}
where $C(i)$ is the set of counterfactual samples for the i-th individual of the test set. This corresponds to counterfactuals sampled with the Adversarial inference process defined at step 1 (with the best configuration reported in tables \ref{tab:binary_case} and \ref{tab:continuous_case}). As discussed above, the synthetic datasets allow one to rely on "true" counterfactuals for the computation of counterfactual fairness, rather than relying on an inference process which may include some bias. For these datasets, we thus also report an additional RealCF metric, which is defined as in Eq. \ref{cf}, but using these counterfactuals sampled from the true codes used to generate the test data. For both CF and RealCF, for every $i$ from the test set, $|C(i)|$ equals 1 for binary settings and $|C(i)|$ equals 1000 for the continuous one.  $\Delta$ is a cost function between two predictions, the  logit paring cost for the binary case (more details  given in section \ref{counterfact} step 2) and a simple squared difference for the continuous setting. 

Results from both tables first confirm the good behavior of our inference model from step 1, which allows one to obtain greatly better results than other inference processes for both the discrete and the continuous settings. Our adversarial counterfactual inference framework allows one to get codes that can be easily used to generate relevant counterfactual individuals. For this observation, the most important results are those given for the synthetic scenarios, for which the RealCF metric shows  good results for our method, while strongly  reliable since relying on counterfactuals sampled from true codes of individuals. 

Secondly, results from table \ref{tab:continuous_case} show that, even in the continuous setting where the enumeration of all values from $\Omega_A$ is not possible, it is possible to define counterfactual mitigation methods such as our approaches $L_{CF}$ and $L_{DynCF}$. These two methods, used in conjunction with our  Adversarial Inference, give significantly better results than no mitigation on every dataset. Interestingly, we also observe that $L_{DynCF}$ allows one to improve results over $L_{CF}$, which shows the relevance of the proposed dynamic sampling process.
Furthermore, note that we can reasonably expect even better results compared to $L_{CF}$  on  data  with higher-dimensional sensitive attributes. 

\begin{figure}[h]
  \centering
  \includegraphics[scale=0.55,valign=t]{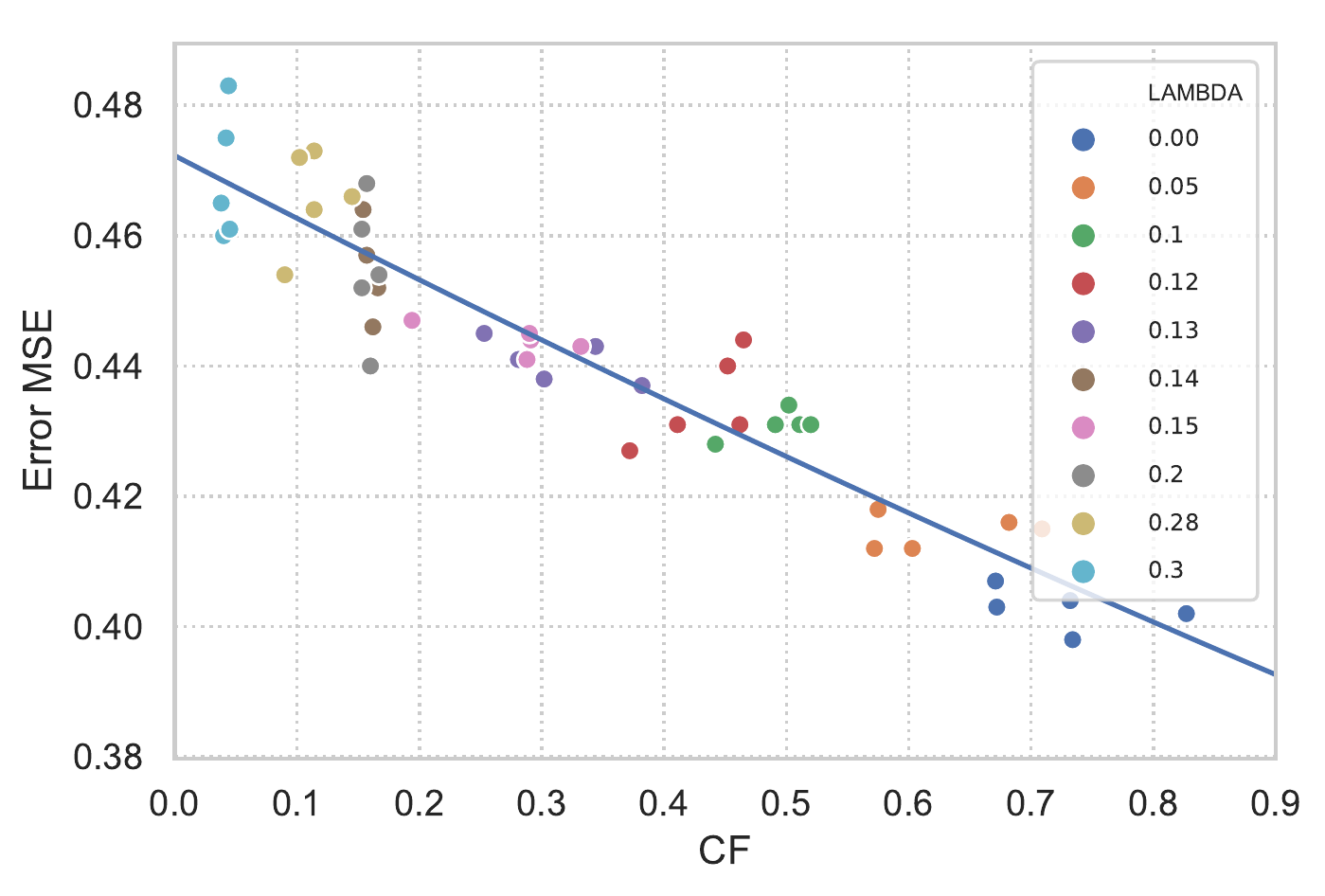}
  \caption{Impact of hyperparameter $\lambda$ (Crime data set)} 
 
  \label{fig:Impact_Hyperparameter}
\end{figure}

To illustrate the impact of the hyperparameter $\lambda$ on the predictions accuracy (MSE Error) and the counterfactual fairness estimation (CF), we plot results for  10 different values of $\lambda$ (5 runs each) on figure \ref{fig:Impact_Hyperparameter} for the Crime data set. 
It clearly confirms that higher values of $\lambda$ produce fairer predictions, while a value near 0 allows one to only focus on optimizing the predictor loss. This is also observable from Fig.  \ref{fig:Error_wrt_A} which plots counterfactual predictions for a specific instance $i$ from the test set.  Higher values of $\lambda$ produce clearly more stable counterfactual predictions. 

\begin{table*}[h!]
\caption{Counterfactual Fairness Results for the Discrete  Case}
\label{tab:binary_case2}
\centering
\resizebox{\textwidth}{!}{
\begin{tabular}{||ll||l|l||l|l||l|l||l|l|l||}
\hhline{===========} 
\multicolumn{1}{||c}{\multirow{2}{*}{Inference}} & \multicolumn{1}{c||}{\multirow{2}{*}{Mitigation}} 
 & \multicolumn{2}{c||}{Adult UCI}      & \multicolumn{2}{c||}{Compas}                                &   \multicolumn{2}{c||}{Bank}    
   &   \multicolumn{3}{c||}{Synthetic Scenario}     
   \\ \cline{3-11}
   \multicolumn{2}{||c||}{ } &  Accuracy & CF 
&  Accuracy & CF
& Accuracy & CF  
& Accuracy & CF  & Real CF  
\\ \hline
\multirow{2}{*}{{Without Constraint}}
& None  
& 84.22\%   &   0.0096
&  67.12\%  &    0.0102
&  90.64\%  &  0.0369  
&  99.49\%  & 0.1087   & 0.1810
\\
&  $L_{CF}$ 
&    83.28\%   &  0.0008
&  66.20\%  &  0.0051
&  90.46\%   &  0.0024 
&  95.89\% & 0.0757   & 0.1327
\\ \hline
\multirow{2}{*}{{$MMD$}}
&  None  
&  84.22\%  &   0.0116
&  67.12\%  &   0.0076 
&  90.64\%  &    0.0469
&  99.49\% &   0.1074 & 0.1775
\\ 
&  $L_{CF}$ 
&   83.84\% &   0.0024
&   65.91\%  &    0.0041
&   90.64\%  &  0.0043
&  99.29\% &  0.0893 &  0.1557
\\ \hline
\multirow{2}{*}{{Adversarial}}
&  None  
&  84.22\%  & 0.0114
&  67.12\%  & 0.0118
&  90.64\%  & 0.0376
&  99.49\%  & 0.1426  &  0.1838
\\ & $L_{CF}$  
&   83.74\%  &   0.0002
&   66.73\%  &    0.0001
&   90.60\%   &  0.000
&  93.19\%  & 0.0001   &  0.0014

\\ \hline
\hhline{===========} 
\end{tabular}
}
\\
\end{table*}

\begin{table*}[h!]
\caption{Counterfactual Fairness Results for the Continuous Case}
\label{tab:continuous_case2}
\centering
\resizebox{\textwidth}{!}{
\begin{tabular}{||ll||l|l||l|l||l|l||l|l|l||}

\hhline{===========} 
\multicolumn{1}{||c}{\multirow{2}{*}{Inference}} & \multicolumn{1}{c||}{\multirow{2}{*}{Mitigation}} &
\multicolumn{2}{c||}{US Census}      & \multicolumn{2}{c||}{Motor}                                &   \multicolumn{2}{c||}{Crime}    
   &   \multicolumn{3}{c||}{Synthetic Scenario}     
   \\ \cline{3-11}
   \multicolumn{2}{||c||}{ } &  Accuracy & CF 
&  MSE & CF
& MSE & CF  
& MSE & CF  & Real CF  
\\ \hline
\multirow{2}{*}{{Adversarial}}
& None
&   0.274      &    0.0615  
&   0.938      &    0.0285  
&   0.412     &    0.7412     
&     0.454    & 0.2490 & 1.1248
\\
& $L_{CF}$ 
 &      0.289      &    0.0009      
 &         0.941    &    0.0009   
 &         0.452    &    0.0154    
 & 0.572  &  0.0014 & 0.2013
\\
&  $L_{DynCF}$  
       &    0.290      &  0.0008   &            0.940          &       0.0005      &    0.445      &     0.0076  &
0.568   &  0.0013   & 0.2000 
\\ \hline
\multirow{2}{*}{{Without Constraint}}
& None
&     0.274    &  0.0433 
&     0.938  &  0.0271
&     0.381  &  0.7219            
&  0.454     &   0.2919   & 1.1338
\\
 & $L_{CF}$  
    &         0.307          &     0.0010                 &   0.939         &         0.0021                &     0.407       &                  0.2938  &  0.531 & 0.1968 & 0.3303
\\ 
& $L_{DynCF}$         &    0.310      &     0.0008  &       0.942        &  0.0016       &    0.418   &    0.2881  & 0.546 & 0.1743 & 0.3188
\\ \hline
\hhline{===========} 
\end{tabular}
}
\\
\end{table*}


\begin{figure}[h!]
\centering
\includegraphics[scale=0.45,valign=t]{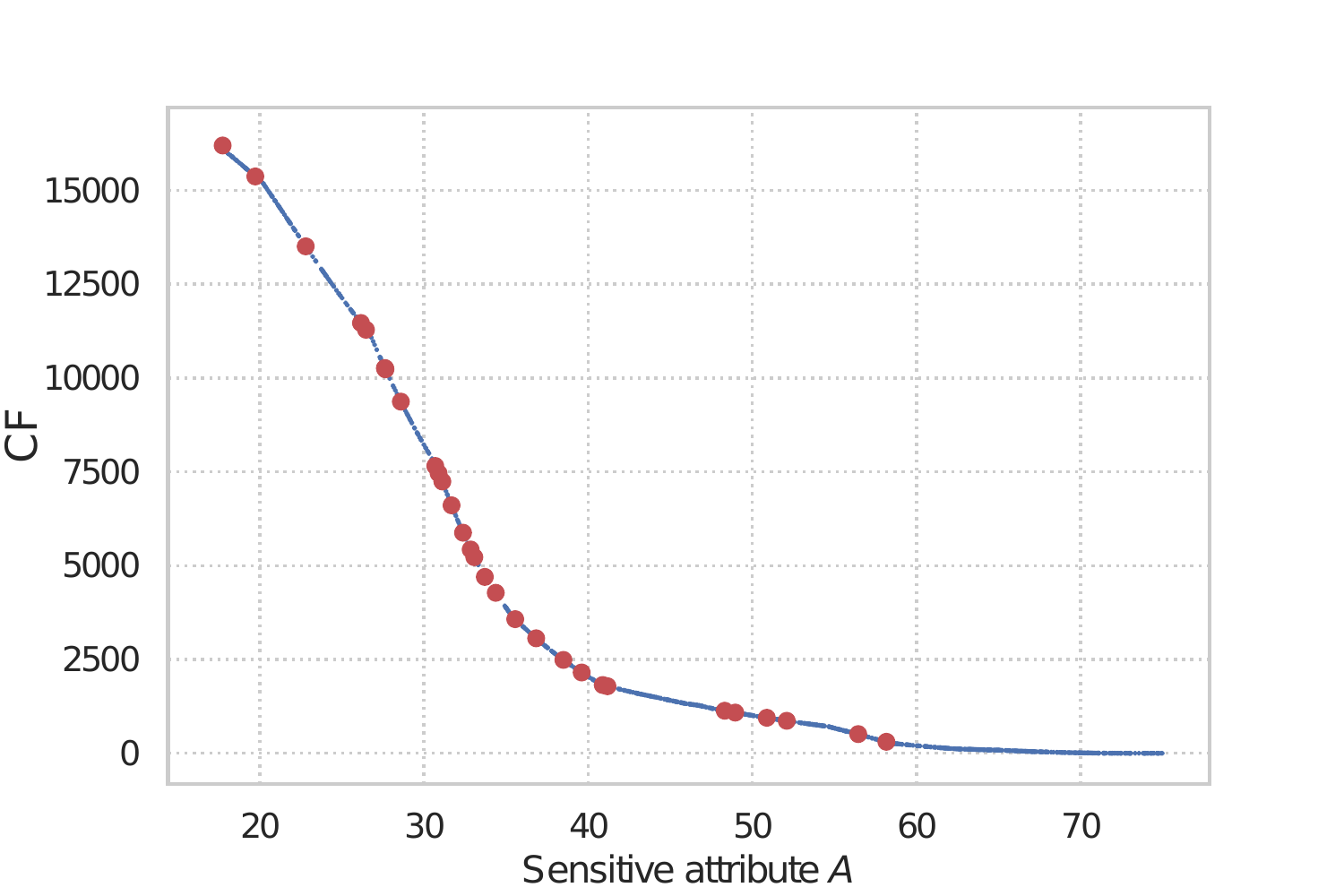}
\caption{Dynamic Sampling Visualization for a randomly sampled individual whose age A is $75$. Red points are sampled counterfactuals  from the dynamic distribution $P_\phi(a'|u)$ with $u$ the inferred confounding for this individual.}
\label{fig:dynSample}
\end{figure}

In figure \ref{fig:dynSample}, we consider the distribution of considered counterfactual samples w.r.t. to the sensitive variable A for the uniform sampling strategy from $P'(A)$ and the dynamic strategy as defined in Eq.\ref{dyn}. This is done on the 
Motor dataset and 
for a specific randomly sampled instance $i$ with sensitive attribute $a_i=75$, at a given point of the optimization, far before convergence (the model is clearly unfair at this point). The blue points are the counterfactual fairness estimation $(h_\theta({X_{i,A\leftarrow{a}},a)}-h_\theta(X_{i,A\leftarrow{a'}},a'))$ for each counterfactual sampled a' s (1.000 points) from the uniform  distribution $P'(A)$. The red points are the counterfactual fairness estimations for  counterfactuals corresponding to a' values (30 points)  sampled from  our dynamic distribution $P_\phi(a'|u)=\mathcal{N}(\mu_{\phi}(u),\sigma_{\phi}^2(u)I)$, where $\phi$ are the parameters of the adversarial network which optimizes the best mean and variance for each latent code $u$ ($\mu_{\phi}(u)$ and $\sigma_{\phi}^2(u)$). Being optimized to maximize the error at each gradient step, the red points are sampled on lower values of $A$ where the error is the most important. More importantly, very few points are sampled in the easy area, near the true sensitive value of i which is 75. This demonstrates the good behavior of our dynamic sampling process. 

\section{Conclusion}
We  developed  a  new  adversarial  learning  approach  for counterfactual fairness. To the best of our knowledge, this is the first such method that can be applied for  continuous sensitive attributes. 
The method proved to be very efficient for different dependence metrics on various artificial and real-world data sets, for both the discrete and the continuous settings. 
Finally, our proposal is applicable for any causal graph to achieve generic counterfactual fairness. As future works, it might be interesting to consider a generalization of our proposal 
for Path Specific \citep{chiappa2019path} counterfactual fairness  in the continuous case.

\nocite{*}
\bibliography{main}

\begin{thebibliography}{10}
\providecommand{\url}[1]{#1}
\csname url@samestyle\endcsname
\providecommand{\newblock}{\relax}
\providecommand{\bibinfo}[2]{#2}
\providecommand{\BIBentrySTDinterwordspacing}{\spaceskip=0pt\relax}
\providecommand{\BIBentryALTinterwordstretchfactor}{4}
\providecommand{\BIBentryALTinterwordspacing}{\spaceskip=\fontdimen2\font plus
\BIBentryALTinterwordstretchfactor\fontdimen3\font minus
  \fontdimen4\font\relax}
\providecommand{\BIBforeignlanguage}[2]{{%
\expandafter\ifx\csname l@#1\endcsname\relax
\typeout{** WARNING: IEEEtran.bst: No hyphenation pattern has been}%
\typeout{** loaded for the language `#1'. Using the pattern for}%
\typeout{** the default language instead.}%
\else
\language=\csname l@#1\endcsname
\fi
#2}}
\providecommand{\BIBdecl}{\relax}
\BIBdecl

\bibitem{bolukbasi2016man}
T.~Bolukbasi, K.-W. Chang, J.~Y. Zou, V.~Saligrama, and A.~T. Kalai, ``Man is
  to computer programmer as woman is to homemaker? debiasing word embeddings,''
  in \emph{Advances in neural information processing systems}, 2016, pp.
  4349--4357.

\bibitem{angwin2016machine}
J.~Angwin, J.~Larson, S.~Mattu, and L.~Kirchner, ``{Machine bias. ProPublica,
  May 23, 2016},'' 2016.

\bibitem{pedreshi2008discrimination}
D.~Pedreshi, S.~Ruggieri, and F.~Turini, ``Discrimination-aware data mining,''
  in \emph{KDD'08}, 2008, pp. 560--568.

\bibitem{calders2009building}
T.~Calders, F.~Kamiran, and M.~Pechenizkiy, ``Building classifiers with
  independency constraints,'' in \emph{ICDM Workshops}.\hskip 1em plus 0.5em
  minus 0.4em\relax IEEE, 2009, pp. 13--18.

\bibitem{zafar2015fairness}
M.~B. Zafar, I.~Valera, M.~G. Rodriguez, and K.~P. Gummadi, ``Fairness
  constraints: Mechanisms for fair classification,'' \emph{arXiv preprint
  arXiv:1507.05259}, 2015.

\bibitem{kusner2017counterfactual}
M.~J. Kusner, J.~Loftus, C.~Russell, and R.~Silva, ``Counterfactual fairness,''
  in \emph{Advances in Neural Information Processing Systems}, 2017, pp.
  4066--4076.

\bibitem{chiappa2019path}
S.~Chiappa, ``Path-specific counterfactual fairness,'' in \emph{Proceedings of
  the AAAI Conference on Artificial Intelligence}, vol.~33, 2019, pp.
  7801--7808.

\bibitem{hinnefeld2018evaluating}
J.~H. Hinnefeld, P.~Cooman, N.~Mammo, and R.~Deese, ``Evaluating fairness
  metrics in the presence of dataset bias,'' \emph{arXiv preprint
  arXiv:1809.09245}, 2018.

\bibitem{hardt2016equality}
M.~Hardt, E.~Price, and N.~Srebro, ``Equality of opportunity in supervised
  learning,'' in \emph{Advances in neural information processing systems},
  2016, pp. 3315--3323.

\bibitem{dwork2012fairness}
C.~Dwork, M.~Hardt, T.~Pitassi, O.~Reingold, and R.~Zemel, ``Fairness through
  awareness,'' in \emph{ITCS'12}, 2012, pp. 214--226.

\bibitem{zhang2018mitigating}
B.~H. Zhang, B.~Lemoine, and M.~Mitchell, ``Mitigating unwanted biases with
  adversarial learning,'' in \emph{AAAI'18}, 2018, pp. 335--340.

\bibitem{abs-1911-04929}
V.~Grari, B.~Ruf, S.~Lamprier, and M.~Detyniecki, ``Fairness-aware neural
  r{\'{e}}nyi minimization for continuous features,'' \emph{arXiv:1911.04929},
  2019.

\bibitem{kamiran2012data}
F.~Kamiran and T.~Calders, ``Data preprocessing techniques for classification
  without discrimination,'' \emph{Knowledge and Informatoin Systems}, vol.~33,
  no.~1, pp. 1--33, 2012.

\bibitem{bellamy2018ai}
R.~K. Bellamy, K.~Dey, M.~Hind, S.~C. Hoffman, S.~Houde, K.~Kannan, P.~Lohia,
  J.~Martino, S.~Mehta, A.~Mojsilovic \emph{et~al.}, ``Ai fairness 360: An
  extensible toolkit for detecting, understanding, and mitigating unwanted
  algorithmic bias,'' \emph{arXiv preprint arXiv:1810.01943}, 2018.

\bibitem{calmon2017optimized}
F.~P. Calmon, D.~Wei, K.~N. Ramamurthy, and K.~R. Varshney, ``Optimized data
  pre-processing for discrimination prevention,'' \emph{arXiv preprint
  arXiv:1704.03354}, 2017.

\bibitem{celis2019classification}
L.~E. Celis, L.~Huang, V.~Keswani, and N.~K. Vishnoi, ``Classification with
  fairness constraints: A meta-algorithm with provable guarantees,'' in
  \emph{Proceedings of the Conference on Fairness, Accountability, and
  Transparency}, 2019, pp. 319--328.

\bibitem{wadsworth2018achieving}
C.~Wadsworth, F.~Vera, and C.~Piech, ``Achieving fairness through adversarial
  learning: an application to recidivism prediction,'' \emph{arXiv:1807.00199},
  2018.

\bibitem{louppe2017learning}
G.~Louppe, M.~Kagan, and K.~Cranmer, ``Learning to pivot with adversarial
  networks,'' in \emph{Advances in neural information processing systems},
  2017, pp. 981--990.

\bibitem{chen2019fairness}
J.~Chen, N.~Kallus, X.~Mao, G.~Svacha, and M.~Udell, ``Fairness under
  unawareness: Assessing disparity when protected class is unobserved,'' in
  \emph{Proceedings of the Conference on Fairness, Accountability, and
  Transparency}, 2019, pp. 339--348.

\bibitem{kearns2017preventing}
M.~Kearns, S.~Neel, A.~Roth, and Z.~S. Wu, ``Preventing fairness
  gerrymandering: Auditing and learning for subgroup fairness,'' \emph{arXiv
  preprint arXiv:1711.05144}, 2017.

\bibitem{DBLP:conf/icdm/GrariRLD19}
V.~Grari, B.~Ruf, S.~Lamprier, and M.~Detyniecki, ``Fair adversarial gradient
  tree boosting,'' in \emph{ICDM'19}, 2019, pp. 1060--1065.

\bibitem{adel2019one}
T.~Adel, I.~Valera, Z.~Ghahramani, and A.~Weller, ``One-network adversarial
  fairness,'' in \emph{AAAI'19}, vol.~33, 2019, pp. 2412--2420.

\bibitem{pmlr-v97-mary19a}
J.~Mary, C.~Calauz{\`e}nes, and N.~E. Karoui, ``Fairness-aware learning for
  continuous attributes and treatments,'' in \emph{ICML'19}, 2019, pp.
  4382--4391.

\bibitem{pearl2009causal}
J.~Pearl \emph{et~al.}, ``Causal inference in statistics: An overview,''
  \emph{Statistics surveys}, vol.~3, pp. 96--146, 2009.

\bibitem{madras2019fairness}
D.~Madras, E.~Creager, T.~Pitassi, and R.~Zemel, ``Fairness through causal
  awareness: Learning causal latent-variable models for biased data,'' in
  \emph{Proceedings of the Conference on Fairness, Accountability, and
  Transparency}, 2019, pp. 349--358.

\bibitem{pfohl2019counterfactual}
S.~Pfohl, T.~Duan, D.~Y. Ding, and N.~H. Shah, ``Counterfactual reasoning for
  fair clinical risk prediction,'' \emph{arXiv preprint arXiv:1907.06260},
  2019.

\bibitem{louizos2017causal}
C.~Louizos, U.~Shalit, J.~M. Mooij, D.~Sontag, R.~Zemel, and M.~Welling,
  ``Causal effect inference with deep latent-variable models,'' in
  \emph{Advances in Neural Information Processing Systems}, 2017, pp.
  6446--6456.

\bibitem{russell2017worlds}
C.~Russell, M.~J. Kusner, J.~Loftus, and R.~Silva, ``When worlds collide:
  integrating different counterfactual assumptions in fairness,'' in
  \emph{Advances in Neural Information Processing Systems}, 2017, pp.
  6414--6423.

\bibitem{team2016rstan}
S.~D. Team \emph{et~al.}, ``Rstan: the r interface to stan,'' \emph{R package
  version}, vol.~2, no.~1, 2016.

\bibitem{kingma2013auto}
D.~P. Kingma and M.~Welling, ``Auto-encoding variational bayes,'' \emph{arXiv
  preprint arXiv:1312.6114}, 2013.

\bibitem{shalit2017estimating}
U.~Shalit, F.~D. Johansson, and D.~Sontag, ``Estimating individual treatment
  effect: generalization bounds and algorithms,'' in \emph{ICML'17}, 2017, pp.
  3076--3085.

\bibitem{gretton2012kernel}
A.~Gretton, K.~M. Borgwardt, M.~J. Rasch, B.~Sch{\"o}lkopf, and A.~Smola, ``A
  kernel two-sample test,'' \emph{Journal of Machine Learning Research},
  vol.~13, no. Mar, pp. 723--773, 2012.

\bibitem{zhao2017infovae}
S.~Zhao, J.~Song, and S.~Ermon, ``Infovae: Information maximizing variational
  autoencoders,'' \emph{arXiv preprint arXiv:1706.02262}, 2017.

\bibitem{chen2016variational}
X.~Chen, D.~P. Kingma, T.~Salimans, Y.~Duan, P.~Dhariwal, J.~Schulman,
  I.~Sutskever, and P.~Abbeel, ``Variational lossy autoencoder,'' \emph{arXiv
  preprint arXiv:1611.02731}, 2016.

\bibitem{bowman2015generating}
S.~R. Bowman, L.~Vilnis, O.~Vinyals, A.~M. Dai, R.~Jozefowicz, and S.~Bengio,
  ``Generating sentences from a continuous space,'' \emph{arXiv:1511.06349},
  2015.

\bibitem{sonderby2016ladder}
C.~K. S{\o}nderby, T.~Raiko, L.~Maal{\o}e, S.~K. S{\o}nderby, and O.~Winther,
  ``Ladder variational autoencoders,'' in \emph{NIPS'16}, 2016, pp. 3738--3746.

\bibitem{MakhzaniSJG15}
\BIBentryALTinterwordspacing
A.~Makhzani, J.~Shlens, N.~Jaitly, and I.~J. Goodfellow, ``Adversarial
  autoencoders,'' \emph{CoRR}, vol. abs/1511.05644, 2015. [Online]. Available:
  \url{http://arxiv.org/abs/1511.05644}
\BIBentrySTDinterwordspacing

\bibitem{Goodfellow2014}
I.~J. Goodfellow, J.~Pouget-Abadie, M.~Mirza, B.~Xu, D.~Warde-Farley, S.~Ozair,
  A.~Courville, and Y.~Bengio, ``{Generative Adversarial Networks},'' pp. 1--9,
  2014.

\bibitem{Dua:2019}
D.~Dua and C.~Graff, ``{UCI} ml repository,''
  \url{http://archive.ics.uci.edu/ml}, 2017.

\bibitem{Moro2014}
S.~Moro, P.~Cortez, and P.~Rita, ``A data-driven approach to predict the
  success of bank telemarketing,'' \emph{Decision Support Systems}, vol.~62, 06
  2014.

\bibitem{USCensus}
{US Census Bureau}, ``Us census demographic data,''
  \url{https://data.census.gov/cedsci/}, online; accessed 03 April 2019.

\bibitem{pricinggame15}
{The Institute of Actuaries of France}, ``Pricing game 2015,''
  \url{https://freakonometrics.hypotheses.org/20191}, online; accessed 14
  August 2019.

\bibitem{louizos2015variational}
C.~Louizos, K.~Swersky, Y.~Li, M.~Welling, and R.~Zemel, ``The variational fair
  autoencoder,'' \emph{arXiv preprint arXiv:1511.00830}, 2015.

\bibitem{pearl2000}
L.~Neuberg, ``{Causality: models, reasoning, and inference, by judea pearl,
  cambridge university press, 2000},'' \emph{Econometric Theory}, vol.~19, pp.
  675--685, 2003.

\bibitem{Yeh:2009:CDM:1464526.1465163}
I.-C. Yeh and C.-h. Lien, ``The comparisons of data mining techniques for the
  predictive accuracy of probability of default of credit card clients,''
  \emph{Expert Syst. Appl.}, vol.~36, no.~2, pp. 2473--2480, Mar. 2009.

\end{thebibliography}
\bibliographystyle{IEEEtran}
\end{document}